\documentclass[conference]{IEEEtran}
\IEEEoverridecommandlockouts
\usepackage{hyperref}
\pdfoutput=1

\usepackage{cite}
\usepackage{amsmath,amssymb,amsfonts}
\usepackage{algorithmic}
\usepackage{graphicx}
\usepackage{url}
\ifCLASSOPTIONcompsoc
    \usepackage[caption=false, font=normalsize, labelfont=sf, textfont=sf]{subfig}
\else
\usepackage[caption=false, font=footnotesize]{subfig}
\fi
\usepackage{textcomp}
\usepackage{xcolor}
\usepackage{mathtools}
\usepackage[utf8]{inputenc}

\def\BibTeX{{\rm B\kern-.05em{\sc i\kern-.025em b}\kern-.08em
    T\kern-.1667em\lower.7ex\hbox{E}\kern-.125emX}}
\begin{document}

\title{Finding the most similar textual documents using Case-Based Reasoning\\
}

\newcommand{\norm}[1]{\left\lVert#1\right\rVert}
\DeclarePairedDelimiter\abs{\lvert}{\rvert}%

\author{\IEEEauthorblockN{\thanks{\IEEEauthorrefmark{1}This work has been done while the author was at Mälardalen University.}Marko Mihajlovic\IEEEauthorrefmark{1}}
\IEEEauthorblockA{\textit{Department of Computer Science} \\
\textit{ETH Zurich}\\
Zurich, Switzerland \\
Email: markomih@ethz.ch}
\and
\IEEEauthorblockN{Ning Xiong}
\IEEEauthorblockA{\textit{School of Innovation, Design and Engineering} \\
\textit{Malardalen University}\\
Västerås, Sweden \\
Email: ning.xiong@mdh.se}
}

% \author{\IEEEauthorblockN{Marko Mihajlović\IEEEauthorrefmark{1},
% % Ermal Bizhuta\IEEEauthorrefmark{1}, Caterina Muntaner\IEEEauthorrefmark{1}, Jesús Miguel Álvarez\IEEEauthorrefmark{1}, \thanks{\IEEEauthorrefmark{2}Supervisor}
% \thanks{\IEEEauthorrefmark{2}Mälardalen University, Västerås, Sweden. ning.xiong@mdh.se}Ning Xiong\IEEEauthorrefmark{2}}}
% % \IEEEauthorblockA{Academy of Innovation, Design, and Engineering,
% % Mälardalen University\\
% % Västerås, Sweden\\
% % Email: \IEEEauthorrefmark{1}markomih@ethz.ch,
% % \IEEEauthorrefmark{2}ning.xiong@mdh.se,
% % }}

\maketitle

\begin{abstract}
In recent years, huge amounts of unstructured textual data on the Internet are a big difficulty for AI algorithms to provide the best recommendations for users and their search queries. Since the Internet became widespread, a lot of research has been done in the field of Natural Language Processing (NLP) and machine learning. Almost every solution transforms documents into Vector Space Models (VSM) in order to apply AI algorithms over them. One such approach is based on Case-Based Reasoning (CBR). Therefore, the most important part of those systems is to compute the similarity between numerical data points. In 2016, the new similarity TS-SS metric is proposed, which showed state-of-the-art results in the field of textual mining for unsupervised learning. However, no one before has investigated its performances for supervised learning (classification task). In this work, we devised a CBR system capable of finding the most similar documents for a given query aiming to investigate performances of the new state-of-the-art metric, TS-SS, in addition to the two other geometrical similarity measures --- Euclidean distance and Cosine similarity --- that showed the best predictive results over several benchmark corpora. The results show surprising inappropriateness of TS-SS measure for high dimensional features. 
\end{abstract} 


\section{Introduction}
Nowadays, almost every person in the World generates data on the Internet; social media, news, public comments, blogs, searching the Internet, etc. All this information is recorded in a database which makes a problem known as Big Data \cite{BigData}. Accordingly, vast amounts of data are unstructured textual data which makes inconvenience for machine learning models to harness their predictive power. 

Consequently, diverse approaches and algorithms have been proposed to deal with this issue. However, most of them work well only with numerical data. To use those algorithms, we need to convert textual data into numerical feature vectors. Then, AI algorithms are fed with those data in order to learn the important features that lead the final successful prediction. But first, before the whole system is implemented, we need a textual corpus to show whether our approach is better than the available ones. After the data is acquired, it needs to be preprocessed, which means that the original text is altered in a way to be more suitable for further use --- reducing vocabulary, outliers, and other impurities. Such a modified text is then ready for feature engineering. This process uses statistics to construct numerical feature vectors from raw textual data. In data mining and information retrieval this procedure of converting text into a numerical vector is known as Vector Space Model (VSM) \cite{VSM}; a set of linearly independent basis vectors that represent textual documents. Afterwards, popular approaches, such as Case-Based Reasoning (CBR) \cite{CBR}, can be used to find the most similar documents based on already seen cases (training data). The system’s recommendation for the new document can be evaluated by an expert and added to the pool of training data in order to further improve the system’s predictive performances. However, to harness the power of CBR system, we need to construct a similarity metric that can capture the important characteristics of feature vectors. 

The major contribution of this work is investigation of performances of the similarity metric TS-SS (Triangle’s area Similarity - Sector’s area Similarity) \cite{TSSS}, proposed by Heidarian and Dinneen, that has shown state-of-the-art performances for document clustering. The results of this algorithms are compared with the other well-known similarity measures, Euclidean Distance (ED) and Cosine Similarity (CS). Also, we provide a theoretical justification why TS-SS measure is incapable of capturing feature differences among data points.

Section II discusses state-of-the-art methods and related work done in the field of information retrieval and data mining for each of the stages in the process. Section III describes an implementation of our system and methods for performance evaluation. Section IV has performance measurements of our implementation for a variety of similarity metrics. Section V explores the reasons and provides theoretical justification for achieved performances. Finally, a short summary of our research and future work is given in Section VI.

The code for this project is publicly available\footnote{\url{https://github.com/Maki94/document-classification}}.

\section{Related Work}
Many systems have been devised to overcome difficulties with recognizing the most similar textual documents. This process includes several independent steps, and a lot of research has been done in each of these steps to improve systems' predictive performances. 

% \subsection*{WORK SPLIT}
% Ermal Bizhuta wrote the introduction and the part about standardized datasets.
% Marko Mihajlovic wrote data preprocessing part and discussion and results section. 
% Caterina Muntaner wrote similarity metrics section. 
% Jesús Miguel Álvarez wrote feature engineering section. 

\subsection*{STANDARDIZED DATASETS}
The first step required by any machine learning system is to find a benchmark dataset for performance evaluation. According to Larson, standard test collections for information retrieval are \textit{The Cranfield collection}, \textit{NTCIR}, \textit{Reuters Corpus}, \textit{20 NewsGroups}, and several other corpora \cite{IRBOOK}. 

\subsection*{DATA PREPROCESSING}
Before the features are extracted from the raw textual data, the textual data should be altered in order to reduce the vocabulary size and decrease inaccuracies in feature representation. This process can be briefly divided into 3 categories --- dropping specific terms, word replacement, and stemming.

Some specific words do not bring any value, and they should be excluded from the dataset. This procedure mainly depends on textual corpora; for example, if the data includes web pages, then HTML tags should be removed, if it contains XML files, then XML tags should be eliminated, etc. Additionally, some words are contained only in a couple of documents, which make them too specific to be used in overall system. On the other side, some words too frequently occur in every document and they should be removed as well; for example, if all documents are about computer science, then term computer is irrelevant, and should be excluded from feature space. In our scenario, we are going to focus on regular textual data, so we will not further examine specific cases as with HTML and XML. Larson \cite{IRBOOK} recommends some procedures that are considered as a good practice --- removing stop words, eliminating punctuation, making the word lowercase, and many other procedures which are described in detail in the implementation phase.

The purpose of the word replacement procedure is to reduce the vocabulary size. This process includes spelling correction, synonym replacement, and specific replacements. For spelling correction, it is widely used edit distance based on Levenshtein distance \cite{Ldistance} to find a well spelled word. State-of-the-art approach for synonym replacement is based on WordNet \cite{WordNet}. Other corrections include simple word concatenation, number mapping and other procedures to overcome dataset peculiarities; for example, mapping mac book to \textit{macbook}, every number to the one token, etc.  
 
The next process is stemming – replacing each words with its base form. The Porter stemming algorithm \cite{Porter} has been recommended by most authors for natural language processing tasks. For example, by applying the Porter stemmer, the word women will be replaced by woman, plays by play, etc. 

\subsection*{FEATURE ENGINEERING}
Several successful feature extraction methods for NLP tasks have been proposed and improved over the past decades. These methods map words or phrases from the vocabulary to vectors of real numbers. The most popular approaches for constructing feature vectors are: bag-of-words model (BoW) \cite{BoW}, tf-idf \cite{tfidf}, Glove \cite{Glove}, and word2vec \cite{word2vec}.

BoW method simply counts each word and its number of occurrences is recorded in feature vector at a specific position for that term. One obvious drawback of this approach is that it favors longer documents, therefore, tf-idf measure was introduced to overcome this problem by calculating product between term frequencies and inverse document frequency. Now inverse document frequency will decrease bias towards longer documents. Glove method also counts how frequently a word appears in a context/document, but it uses dimension reduction techniques to achieve low-dimension representation, hence feature vectors lose interpretability, whereas word2vec suffer from the same drawback. It is constructed by a neural network, which results in representing words/phrases as their probability distribution.  

\subsection*{SIMILARITY METRICS}
To find the most similar document, numerical feature vectors should be compared by calculating a similarity metric. A lot of metrics have been invented to capture the most important features of a vector. Those metrics can be divided into two subcategories: geometrical and non-geometrical methods. Summary of these approaches can be found in \cite{TSSS}. 

\subsection*{EVALUATION METHODS}
A standard way of evaluating the quality of classification algorithms is based on confusion matrix, and derived measures from it, such as accuracy, precision, recall, and $F_b$ score.   

\section{METHODS AND IMPLEMENTATION}
\begin{figure}[htbp]
\centerline{\includegraphics[scale=.65]{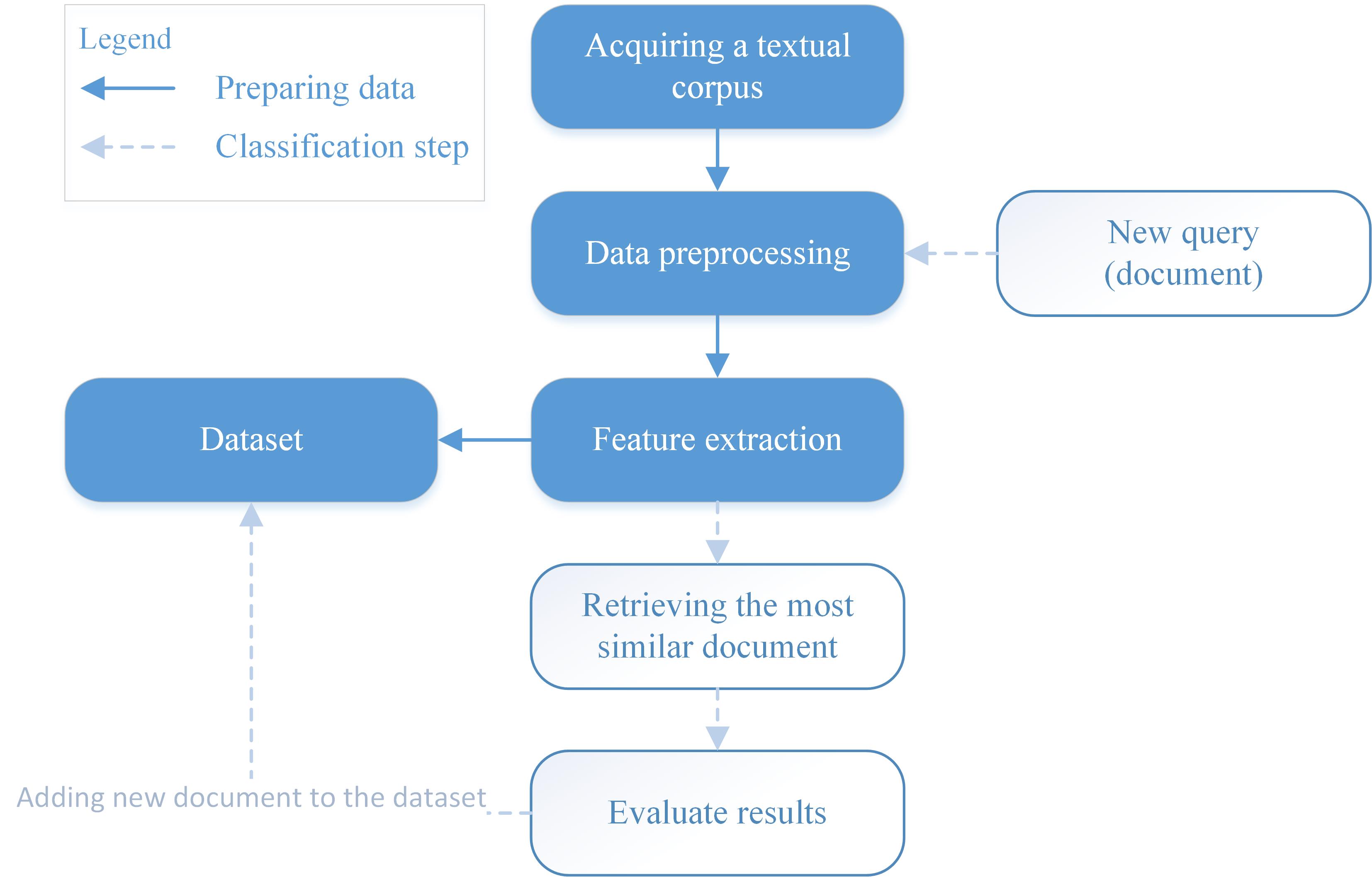}}
\caption{Architecture of the implemented system}
\label{fig:flow-diagram}
\end{figure}

Based on the previous work, the architecture of our system is devised accordingly. \figurename~\ref{fig:flow-diagram} depicts modular architecture of our system. 

The first phase of our system is to acquire some training data, which will be used for searching similar documents. Those documents are then filtered by a preprocessing procedure in order to get rid of outliers and reduce the size of vocabulary. The next phase is to extract features from modified textual documents and to save them in order to accelerate the classification step. When the features are extracted, the most similar document is retrieved for a new query (document). The similarity between two vectors is calculated by several similarity measures. The outcome for each query is recorded and reviewed by an expert, then, the new document can be added to the pool of training documents, which will lead to the improvement in predictive performance for the future queries.  

\subsection*{STANDARDIZED DATASETS}
The first step is to acquire some data. We used five different datasets. 20 NewsGroups\footnote{\url{scikit-learn.org}} dataset which comprises of 18864 newsgroups posts on 20 topics, Reuters\footnote{\url{nltk.org}} dataset which contains 10,788 news documents on 90 topics, the first million-word electronic corpus of English, created in 1961 at Brown University --- Brown dataset\footnotemark[3] --- each document is categorized by one out of fifteen genres. The other two datasets --- the Movie Review\footnotemark[3] and Sentence Polarity\footnotemark[3] datasets --- are labeled with binary values positive or negative. The reason for our choice is that these corpora are different in the number of documents and labels, which can reveal different characteristics of used similarity metrics. Both the datasets are split into training and test dataset. 

\begin{table}
\centering
\caption{Feature extraction parameters}
\label{tab:features}
\begin{tabular}{|c|c|}
    \hline
    Parameters & Value \\
    \hline
    Minimum word length & 3 \\ 
    %\hline
    Maximum document frequency & 50\% \\
    %\hline
    Minimum document frequency & 1\%\\
    %\hline
    Lowercase & True\\
    %\hline
    Stop words & English\\
    %\hline
    Analyzer & Only words \\
    %\hline
    Feature extraction & tf-idf\\
    %\hline
    Feature vector dimensionality & Different values are evaluated\\
    \hline
\end{tabular}
\end{table}

\subsection*{DATA PREPROCESSING}
After the data is acquired, it should be altered in order to reduce the size of vocabulary used for feature extraction. Different parameters for the procedures of altering the textual documents are summarized in the Table~\ref{tab:features}. 

The performances of our system deeply depend on the data preparation procedure. First, words that are shorter of three characters are eliminated. Then, based on the term frequency in documents, terms are kept or discarded. Those that are present too frequently, appear in 50\% of the total amount of documents, and those that are too specific, 1\% appearance, are eliminated. After this procedure, all characters are lowercase, and words that do not contain any valuable information, stop words, are removed. 

In the next step, by observing the textual documents it is concluded that specific words need to be eliminated, in our example email and web addresses are also removed. The final step was to discard every character that is not a letter and to apply Porter stemmer to simplify word forms. It can be noted that some recommended procedures, such as mapping numbers to a specific token, are not implemented because achieving the best performances was not an aim of the project. The goal is to  explore the behavior of different similarity metrics when finding the most similar documents. 

\subsection*{FEATURE ENGINEERING}
The data preprocessing procedure has reduced the size of vocabulary significantly. Now, the feature extraction method should convert each document into a numerical feature vector.

Vector space model based on \textit{tf-idf} method usually outperforms other methods with the smaller amount of data. 

For this purpose, \textit{tf-idf} procedure is applied over the given training dataset. To evaluate the performances of different similarity metrics, different lengths of feature vectors are considered. Thus, forcing feature vectors to be of a fixed dimensionality. This is done by removing features with the smallest values. The aim of this experiment is to show how the system behaves in high-dimension space. 

\subsection*{SIMILARITY METRICS}
To retrieve the most similar document, a similarity between two documents needs to be calculated. Each of the documents is first converted to a numerical feature vector, then the similarity is calculated. In this work, we used three similarity measures --– ED (\ref{eq:ed}), CS (\ref{eq:cs}), and TS-SS (\ref{eq:tsss}) --- that showed state-of-the-art performances in many NLP tasks. All these metrics have useful geometrical representation and a short summary of drawbacks and advantages of these methods is well described in \cite{TSSS}. 

\begin{figure}[b]
    \centering
  \subfloat[]{%
       \includegraphics[width=0.5\linewidth]{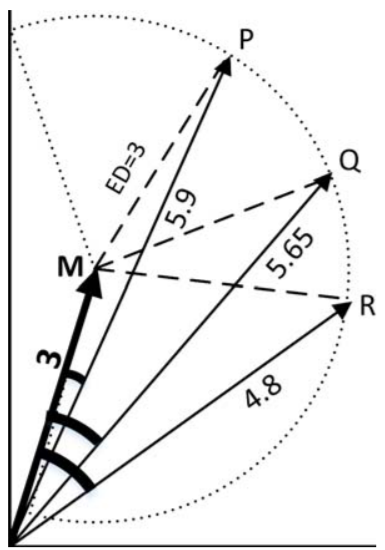}}
    \label{fig:ed}\hfill
  \subfloat[]{%
        \includegraphics[width=0.5\linewidth]{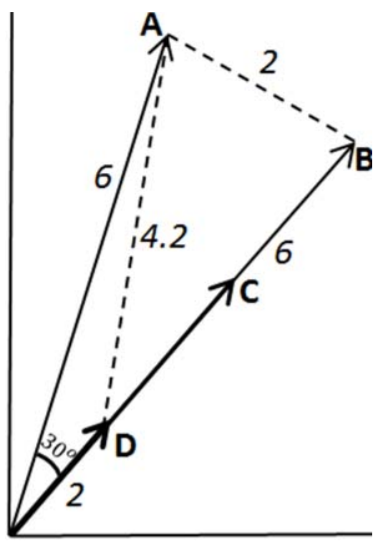}}
    \label{fig:cs}\\
  \caption{(a) Example of Euclidean distance drawback. (b) Example of Cosine similarity drawback.}
  \label{fig:ed_cs} 
\end{figure}
\begin{figure} 
    \centering
  \subfloat[]{%
       \includegraphics[width=0.45\linewidth]{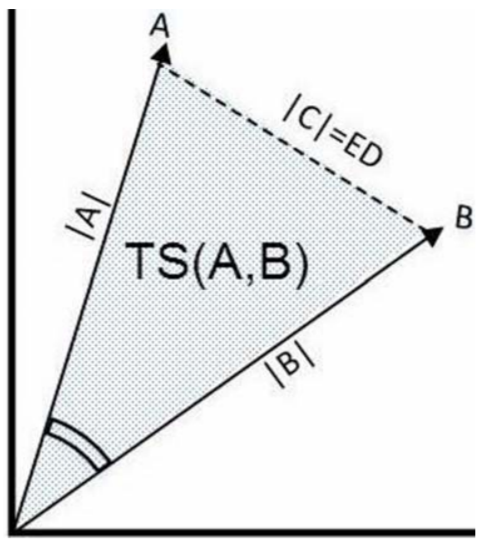}}
    \label{fig:ts}\hfill
  \subfloat[]{%
        \includegraphics[width=0.45\linewidth]{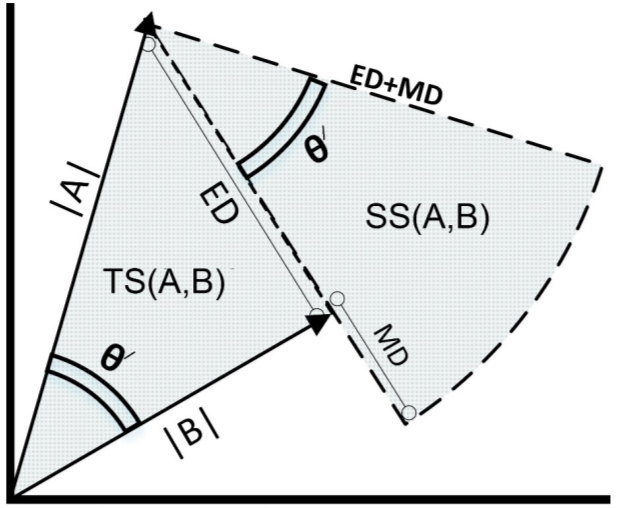}}
    \label{fig:tsss}\\
  \caption{(a) Triangle Similarity (TS). Triangle Similarity---Section Similarity (TS-SS).}
  \label{fig:ts_ss} 
\end{figure}

\begin{equation} \label{eq:ed}
    ED(x_i,x_j)=\sqrt{\sum_{k=1}^M (x_{ik}-x_{jk)^2}}
\end{equation} 
\begin{equation} \label{eq:cs}
    CS(x_i,x_j)=\frac{\sum_{k=1}^M x_{ik}x_{jk}}{\norm{x_i}\norm{x_j}}
\end{equation}
\begin{equation} \label{eq:ts}
    TS(x_i,x_j)=\frac{\norm{x_i}\norm{x_j}\sin{\theta'}}{2}
\end{equation}
\begin{equation} \label{eq:ss}
    SS(x_i,x_j)=\pi(ED(x_i,x_j)+\abs{\norm{x_i}-\norm{x_j}})^2\frac{\theta'}{360}
\end{equation}

\begin{equation} \label{eq:tsss}
    TS\text{-}SS(x_i,x_j)=TS(x_i,x_j)SS(x_i,x_j)
\end{equation}

\begin{equation} \label{eq:theta}
    \theta'(x_i,x_j)=\arccos{CS(x_i, x_j)} + 10
\end{equation}

The reason for introducing a novel similarity measure, TS-SS, is justified by weaknesses of the Euclidean distance and cosine similarity. The drawback of ED can be illustrated in 2-dimensional space (\figurename~\ref{fig:ed_cs}), it can be clearly seen that $ED(M,P) \simeq ED(M,Q) \simeq ED(M,R)$ holds; however, vectors $P,Q,R$ differ significantly. One clear disadvantage of ED is not taking angle between two vectors into account. 

On the other side, the cosine similarity does not suffer from this drawback because it only considers the angle between two given vectors. However, the problem with the cosine similarity is that it does not consider the magnitude of vectors. \figurename~\ref{fig:ed_cs} illustrates a scenario when three vectors are equally similar despite their obvious dissimilarity. In particular, statement $CS(A,B) \simeq CS(A,C) \simeq CS(A,D)$ holds. 

To address these weaknesses --- vector magnitude for CS and angle between two vectors for ED --- TS-SS metric was proposed. This measure is calculated as a product between Triangle’s Area Similarity (TS) and Sector’s Area Similarity (SS). 
The former is calculated based on the triangular area between two vectors in the Euclidean space, which alone suffers from the same drawback as ED. The latter is calculated as an area of a circular segment, which is describe by a diameter and an angle. The diameter is equal to the difference between the vectors' magnitudes, while the angle is the angle between two documents. Now, a metric defined as a product of TS and SS should perform better than ED and CS separately because it addresses the drawbacks of both approaches.

\subsection*{EVALUATION METHODS}
The performances are evaluated on the test dataset. For each retrieved document the answer is recorded and compared with the solution, in the end, the probability of correct retrieval is calculated --- accuracy. New documents are not added to the pool of training documents because it would be inconvenient to evaluate predictive performances for different parameters.

\section{Results}
The performances of the implemented system deeply depend on the data preprocessing and feature extraction procedure. Those procedures require tuning several parameters that are summarized in the Table~\ref{tab:features}. 

Three similarity metrics --- ED, CS and TS-SS similarity --– are used for finding the most similar document in the training dataset for a given textual query (document). Then, accuracy is calculated to evaluate predictive performances of our system for different similarity metrics. The whole process of extracting features and searching feature space for the most similar ones is repeated with different tuning parameters and over five datasets; Reuters, 20 NewsGroups, Brown, Movie Review, and Sentence Polarity. 

\subsubsection{Reuters}
\begin{figure}
\centerline{\includegraphics[width=0.5\textwidth]{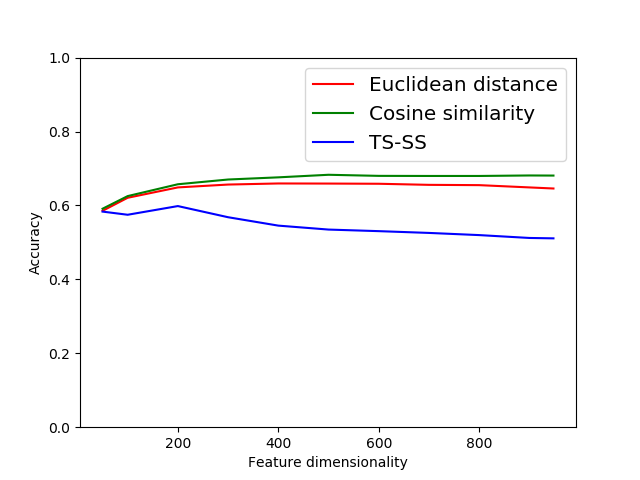}}
\caption{System’s performances over the Reuters dataset without normalized feature vectors}
\label{fig:reuters}
\end{figure}
In the first experiment, the Reuters dataset is used. \figurename~\ref{fig:reuters} shows the accuracy achieved for different dimensionality of the feature vectors. It can be clearly seen that the cosine similarity performed the best, whereas the similarity based on the product of triangle-sector areas was the worse. The method based on the Euclidean distance for the small feature dimensionality follows the same predictive pattern as cosine similarity, but it levels out for the feature length bigger than 300, while the accuracy of cosine similarity constantly increases. In contrast, accuracy of TS-SS similarity fluctuates until the feature dimensionality of 200, after which it constantly decreases.

\subsubsection{20 NewsGroups}
\begin{figure}
\centerline{\includegraphics[width=0.5\textwidth]{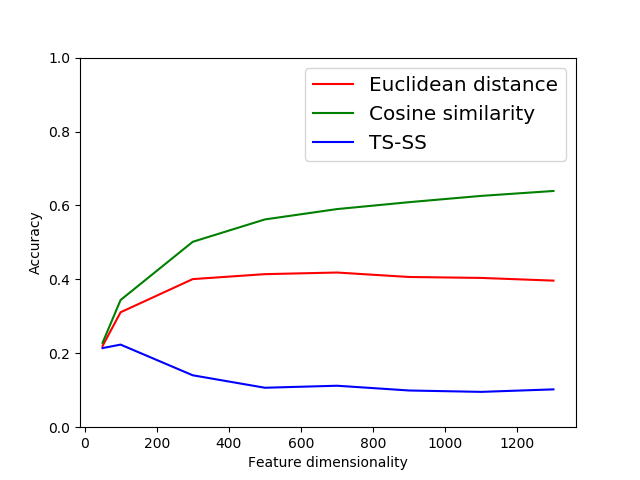}}
\caption{System’s performances over the 20 NewsGroups dataset without normalized feature vectors}
\label{fig:newsgroups}
\end{figure}
For the second experiment, the 20 NewsGroups dataset is used. This dataset has more documents, and so richer vocabulary, which means that feature vectors are of bigger dimensionality; if not limited to a fixed feature length. Therefore, the performances of our system showed a different pattern for every similarity metric. \figurename~\ref{fig:newsgroups} depicts probability that our system will recognize a document’s category successfully for different feature length. In this scenario, the gap between performances of these three metrics is wider. Although the metric based on the cosine angle is still superior to the other ones, it follows logarithmic incline in performances. However, the TS-SS metric, after the slight increase in performances until the feature length of 100, gets worse dramatically until the feature dimensionality of 400, after which it levels out with small oscillations. Euclidean distance’s accuracy, again, shows a similar growth pattern as cosine similarity --- in the beginning it follows cosine similarity until the feature length of 100, after which it levels out, however, the gap between ED's and CS's accuracy is bigger.

\subsubsection{Brown}
\begin{figure}
\centerline{\includegraphics[width=0.5\textwidth]{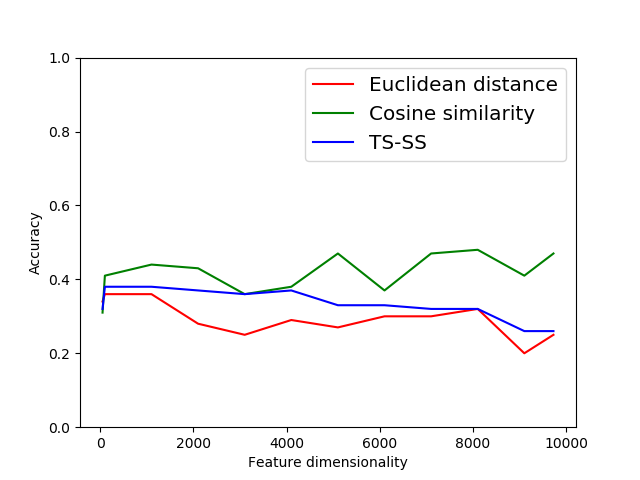}}
\caption{System’s performances over the Brown dataset without normalized feature vectors}
\label{fig:brown}
\end{figure}
The third dataset has fewer categories and feature vectors are of bigger dimensionality. These characteristics of the dataset are opposite from the previous two compared, which help us to reveal different features of similarity metrics. From the \figurename~\ref{fig:brown} it can be that the difference in predictive power between similarity metrics is no longer clear. However, the overall trend is that accuracy decreases with the increase of dimensionality for ED and TS-SS, while CS's accuracy fluctuates and slightly increases over time. One interesting fact is that regardless of the constant fluctuation, CS'S similarity always performs equally or better compared to the other two similarities. 

The next two datasets are binary labeled. 
\subsubsection{Binary datasets}
\begin{figure} 
    \centering
  \subfloat[]{%
       \includegraphics[width=1.0\linewidth]{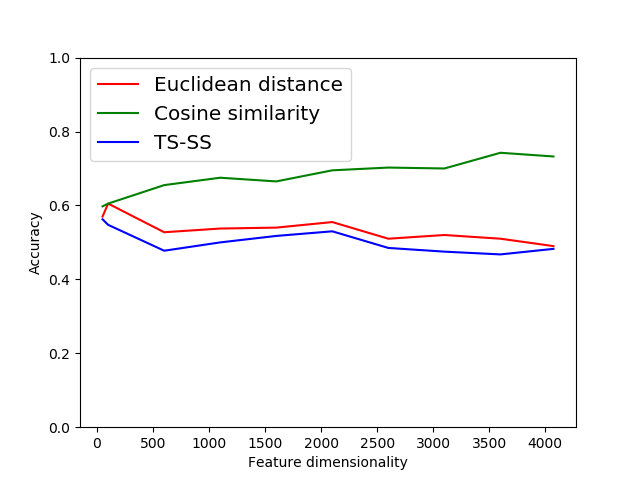}}
    \label{fig:movie_reviews}\\%\hfill
  \subfloat[]{%
        \includegraphics[width=1.0\linewidth]{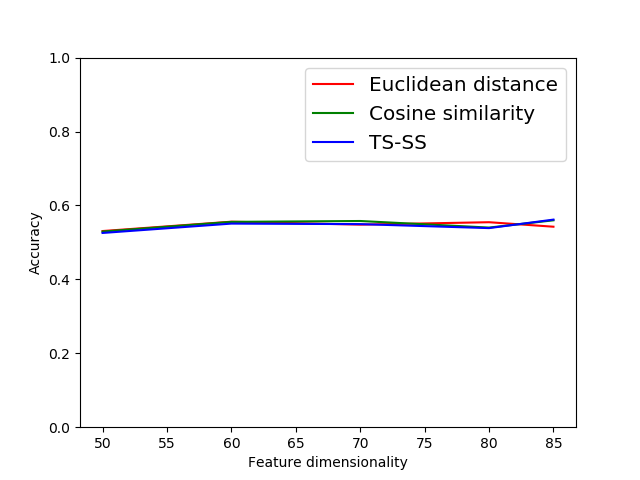}}
    \label{fig:sentence_polarity}\\
  \caption{Predictive performances over two binary datasets: (a) Movie Reviews, and (b) Sentence Polarity.}
  \label{fig:binary_ds} 
\end{figure}
The Movie Reviews dataset has a richer vocabulary, hence feature vectors are of bigger length, while the Sentence Polarity dataset is a smaller one and characterized by less informative features. The \figurename~\ref{fig:binary_ds} clearly demonstrates that for the former dataset cosine similarity is preferred because of its ability to discard mismatched features. On the other side, for the Sentence Polarity dataset the performances of all the similarity metrics are somehow equal, except for the feature length above 75 when TS-SS ability to generate a wider range of values comes into play. 

The cosine similarity shows a constant trend of increase in performances with the incline of feature dimensionality. In other words, the more information a feature vector contains, the better the performances will be. On the other side, the two other methods lack this ability to exploit highly dimensional feature vectors. 

After these five experiments, we further examine the performances of our system when the \textit{tf-idf} feature vectors are normalized. Two procedures are evaluated: normalization based on \textit{l2} and \textit{l1} norm; respectively, each data point $x_i$ is subjected to the constraint (\ref{eq:l2}) and (\ref{eq:l1}), where $M$ is the feature length, parameter $i$ indexes documents, $i \in [1, N]$, and $N$ is the total number of documents in the training dataset. 

\begin{figure}
\centerline{\includegraphics[width=0.5\textwidth]{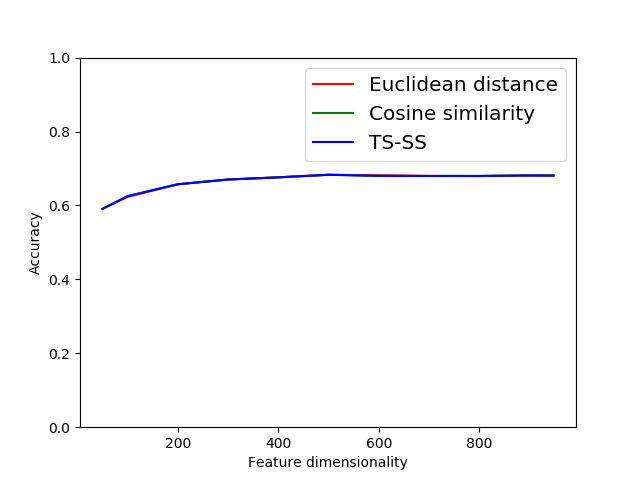}}
\caption{Predictive performances over the Reuters dataset due to l1 (\ref{eq:l1}) constraint.}
\label{fig:reuters_l2}
\end{figure}
\begin{figure}
\centerline{\includegraphics[width=0.5\textwidth]{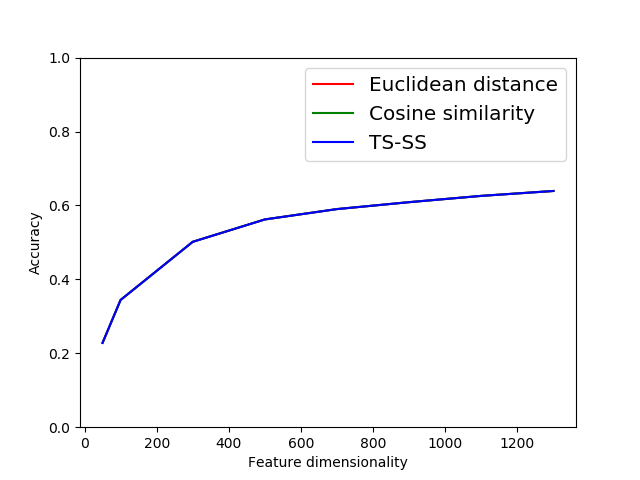}}
\caption{Predictive performances over the 20 NewsGroups dataset due to l2 (\ref{eq:l2}) constraint.}
\label{fig:newsgroups_l2}
\end{figure}
\subsection*{\textit{l2} normalization}
Subjecting feature vectors to the constraint (\ref{eq:l2}), the first five experiments are repeated. Now, the results achieved for these experiments differ significantly. All the three similarity metrics have shown a similar accuracy growth pattern. 

\begin{equation} \label{eq:l2}
    \sqrt{\sum_{k=1}^M x_i^2} = 1
\end{equation}

For the Reuters dataset, the performances of the cosine and TS-SS metrics are almost the same, whereas Euclidean distance is slightly worse for the feature vector length between 100 and 200. From 50 to 250 they show a rapid increase in performances, while after dimensionality of 250 with small oscillations it levels out. For the 20 NewsGroups dataset, the performances of these similarity measures follow the same logarithmic increase in performances. However, the normalization constraint (\ref{eq:l2}) did not impair predictive performances of our system.   

The performances of the other three datasets due to \textit{l2} normalization are shown in the Appendix \ref{Appendix_l2_norm}. However, they all show the same growth pattern as the previous two datasets.

The reason for this phenomenon of showing almost the same growth pattern for each similarity measure is that constraining data points to (\ref{eq:l2}) we get the similar mathematical equations. Now, the similarity metrics are described by mathematical equations (\ref{eq:ed}) for ED, (\ref{eq:cs_l2}) for CS, and (\ref{eq:tsss_l2}) for TS-SS. 

\begin{equation} \label{eq:cs_l2}
    CS(x_i,x_j)=\sum_{k=1}^M x_{ik}x_{jk}
\end{equation}

\begin{equation} \label{eq:tsss_l2}
    TS\text{-}SS(x_i,x_j)=\theta'\sin{\theta'}\textit{ED}(x_i,x_j)
\end{equation}

Note, that fixed scaling parameters that are same for each data point are excluded from the equations because static scaling does not contribute to different document ranking.  

\begin{figure}
\centerline{\includegraphics[width=0.5\textwidth]{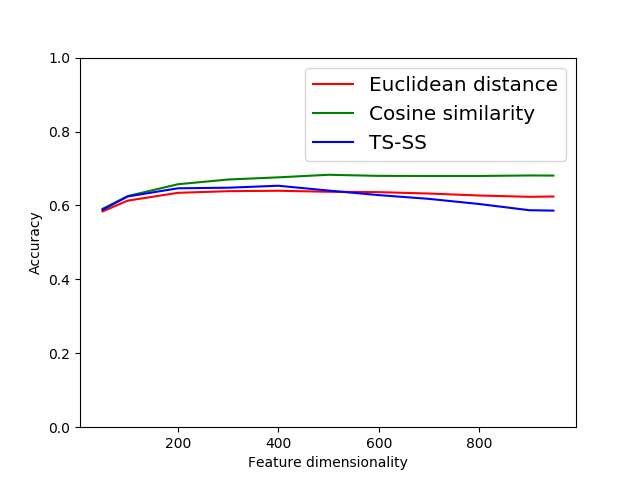}}
\caption{Predictive performances over the Reuters dataset due to l1 (\ref{eq:l1}) constraint.}
\label{fig:reuters_l1}
\end{figure}
\subsection*{\textit{l1} normalization}
When feature vectors are subjected to the constraint \textit{l1} (\ref{eq:l1}), the results are somewhere in between the previous two scenarios. The growth pattern is not nearly the same for each similarity metric as it was with \textit{l2} normalization neither the difference is so drastic as it was without any normalization.  

\begin{equation} \label{eq:l1}
    \sum_{k=1}^M \abs{x_i} = 1
\end{equation}

\figurename~\ref{fig:reuters_l1} depicts the performances of the Reuters dataset due to \textit{l1} constraint, surprisingly unlike in the previous two scenarios, the accuracy of ED and TS-SS do not follow the same pattern for different feature dimensionality. When the dimensionality of the features is smaller TS-SS can capture more details than ED, however with the increase in dimensionality it fails to exhibit this ability. Whereas ED neither benefits of the longer feature factors, but it manages to level out and remain nearly constant.  

In the second experiment \figurename~\ref{fig:reuters_l2}, for the 20 NewsGroup dataset, the pattern of growth for all three similarity metrics is similar to the scenario without any normalization, with the exception that due to \textit{l1} constraint the growth pattern is delayed. Additionally, predictive performances over the other three datasets are virtually the same and are given in the Appendix \ref{Appendix_l1_norm}.

The authors who proposed the TS-SS similarity claim that the features should not be normalized because a normalization constraint would diminish diversity among vectors. However, this characteristic may be beneficial for clustering problems, but for the classification task it is clearly detrimental.

\section{Discussion}
\begin{figure}
\centerline{\includegraphics[width=0.5\textwidth]{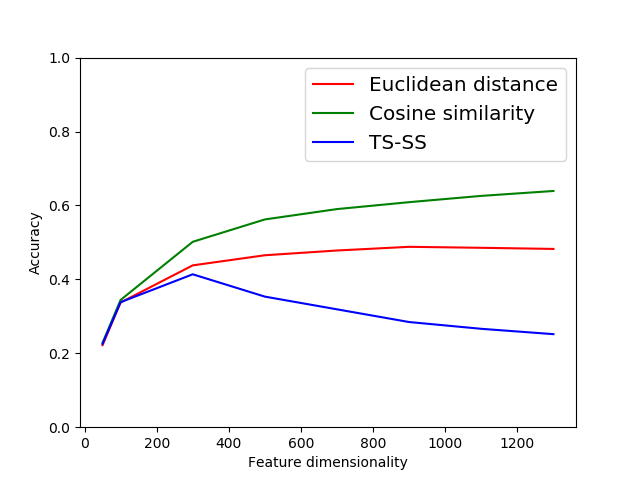}}
\caption{Predictive performances over the 20 NewsGroups dataset due to l1 (\ref{eq:l1}) constraint.}
\label{fig:newsgroups_l1}
\end{figure}

From the results, we can conclude that the metrics based on the Euclidean distance between data points and Triangular similarity suffer from the problem known as \textit{the curse of dimensionality}, which was first introduced by Bellman \cite{bellman1961curse}. 

To exemplify why these measures perform worse as dimensionality increases, consider $N$ data points uniformly distributed in an $M$-dimensional unit ball centered at the origin. Then, our system will classify a given query by finding the closes data point, and assigning it the same class. The median distance from the origin to the closest data point is given by the expression (\ref{eq:closes_point}).

\begin{equation} \label{eq:closes_point}
    \textit{distance}(m,N) = (1-\frac{1}{2}^{\frac{1}{N}})^{\frac{1}{M}}
\end{equation}
From this equation we can derive a formula (\ref{eq:closes_point_N}) to calculate the number of data points required for the given feature length $M$ and expected distance $d$ to the closes data point.

\begin{equation} \label{eq:closes_point_N}
    N = \log_{1-\textit{d}^M} \frac{1}{2}
\end{equation}

Let us fix the expected distance $d$ to the closes data point to the value $0.21$, $d=0.21$, then for $M=1$, we need $3$ data points, for $M=3$, $N=74.5$, and for $M=10$, $N=4155587$. This means that if we want to achieve the same expected accuracy as with the low dimensional feature vectors, we need the size of our training set to grow exponentially. That is why the Euclidean and TS-SS measures show worse performances for high-dimensional feature vectors. 

The authors who proposed TS-SS algorithm, evaluated its performances by four different methods; Uniqueness, Number of Booleans, Minimum gap score, and Purity. However, these methods are more relevant for evaluating document clustering than document classification.

\section{Conclusion and Future Work}

In this work, we performed simple document recommendation based on CBR system in order to evaluate different similarity measures; Euclidean distance, Cosine similarity, and TS-SS similarity. 

We implemented a system that is capable of finding the most similar document in two datasets for a given query (unseen document). Both, the training dataset and queries, are subjected to the same procedures of data preprocessing and feature extraction. Then, three different similarity metrics are employed to retrieve the most similar document, each prediction is recorded so that their performances can be evaluated. After exhaustive testing, we concluded that similarity metric based on cosine outperformed ED and TS-SS in high dimensional space, which leads us to conclusion that ED and TS-SS suffer from a problem known as \textit{the curse of dimensionality}. However, when feature vectors are subjected to the constraint (3), then all three similarity metrics show the similar predictive pattern. 

Recognizing the most similar documents is an active research area in recent years due to increasing use of the Internet. The content on the Internet is mostly unstructured, and for each searched textual document, the similar ones should be retrieved in order to provide the best recommendation to users. Even though a lot of research has been done to improve each step of such a predictive system, there is still no the best solution to deal with NLP problems.  

During our research, we stumbled upon a few interesting research question that we consider worthy of further investigation:

\begin{itemize}
    \item Investigate the performances of similarity metrics for document clustering tasks.
    % \item How the system would perform if feature vectors are constrained to \textit{l1} norm.
    \item Whether the similarity metrics will show different predictive pattern for feature extraction methods such as Glove or word2vec. 
    \item Investigate performances for datasets that have longer textual documents. 
\end{itemize}

\bibliographystyle{IEEEtran.bst}
\bibliography{ms}

\begin{thebibliography}{10}
\providecommand{\url}[1]{#1}
\csname url@samestyle\endcsname
\providecommand{\newblock}{\relax}
\providecommand{\bibinfo}[2]{#2}
\providecommand{\BIBentrySTDinterwordspacing}{\spaceskip=0pt\relax}
\providecommand{\BIBentryALTinterwordstretchfactor}{4}
\providecommand{\BIBentryALTinterwordspacing}{\spaceskip=\fontdimen2\font plus
\BIBentryALTinterwordstretchfactor\fontdimen3\font minus
  \fontdimen4\font\relax}
\providecommand{\BIBforeignlanguage}[2]{{%
\expandafter\ifx\csname l@#1\endcsname\relax
\typeout{** WARNING: IEEEtran.bst: No hyphenation pattern has been}%
\typeout{** loaded for the language `#1'. Using the pattern for}%
\typeout{** the default language instead.}%
\else
\language=\csname l@#1\endcsname
\fi
#2}}
\providecommand{\BIBdecl}{\relax}
\BIBdecl

\bibitem{BigData}
J.~Gantz and D.~Reinsel, ``The digital universe in 2020: Big data, bigger
  digital shadows, and biggest growth in the far east,'' \emph{IDC iView: IDC
  Analyze the future}, vol. 2007, no. 2012, pp. 1--16, 2012.

\bibitem{VSM}
V.~V. Raghavan and S.~M. Wong, ``A critical analysis of vector space model for
  information retrieval,'' \emph{Journal of the American Society for
  information Science}, vol.~37, no.~5, p. 279, 1986.

\bibitem{CBR}
M.~U. Ahmend, S.~Begum, P.~Funk, N.~Xiong, and B.~Von~Sch{\'e}ele, ``Case-based
  reasoning for diagnosis of stress using enhanced cosine and fuzzy
  similarity,'' \emph{Transactions on Case-Based Reasoning for Multimedia
  Data}, vol.~1, no.~1, pp. 3--19, 2008.

\bibitem{TSSS}
A.~Heidarian and M.~J. Dinneen, ``A hybrid geometric approach for measuring
  similarity level among documents and document clustering,'' in \emph{Big Data
  Computing Service and Applications (BigDataService), 2016 IEEE Second
  International Conference on}.\hskip 1em plus 0.5em minus 0.4em\relax IEEE,
  2016, pp. 142--151.

\bibitem{IRBOOK}
R.~R. Larson, ``Introduction to information retrieval,'' \emph{Journal of the
  American Society for Information Science and Technology}, vol.~61, no.~4, pp.
  852--853, 2010.

\bibitem{Ldistance}
L.~Yujian and L.~Bo, ``A normalized levenshtein distance metric,'' \emph{IEEE
  transactions on pattern analysis and machine intelligence}, vol.~29, no.~6,
  pp. 1091--1095, 2007.

\bibitem{WordNet}
G.~A. Miller, ``Wordnet: a lexical database for english,'' \emph{Communications
  of the ACM}, vol.~38, no.~11, pp. 39--41, 1995.

\bibitem{Porter}
P.~Willett, ``The porter stemming algorithm: then and now,'' \emph{Program},
  vol.~40, no.~3, pp. 219--223, 2006.

\bibitem{BoW}
Z.~S. Harris, ``Distributional structure,'' \emph{Word}, vol.~10, no. 2-3, pp.
  146--162, 1954.

\bibitem{tfidf}
K.~Sparck~Jones, ``A statistical interpretation of term specificity and its
  application in retrieval,'' \emph{Journal of documentation}, vol.~28, no.~1,
  pp. 11--21, 1972.

\bibitem{Glove}
J.~Pennington, R.~Socher, and C.~Manning, ``Glove: Global vectors for word
  representation,'' in \emph{Proceedings of the 2014 conference on empirical
  methods in natural language processing (EMNLP)}, 2014, pp. 1532--1543.

\bibitem{word2vec}
T.~Mikolov, I.~Sutskever, K.~Chen, G.~S. Corrado, and J.~Dean, ``Distributed
  representations of words and phrases and their compositionality,'' in
  \emph{Advances in neural information processing systems}, 2013, pp.
  3111--3119.

\bibitem{bellman1961curse}
R.~Bellman, ``Curse of dimensionality,'' \emph{Adaptive control processes: a
  guided tour. Princeton, NJ}, 1961.

\end{thebibliography}

\appendices

\clearpage
\section{}
\label{Appendix_l2_norm}

\begin{figure}[h]
\centerline{\includegraphics[width=0.39\textwidth]{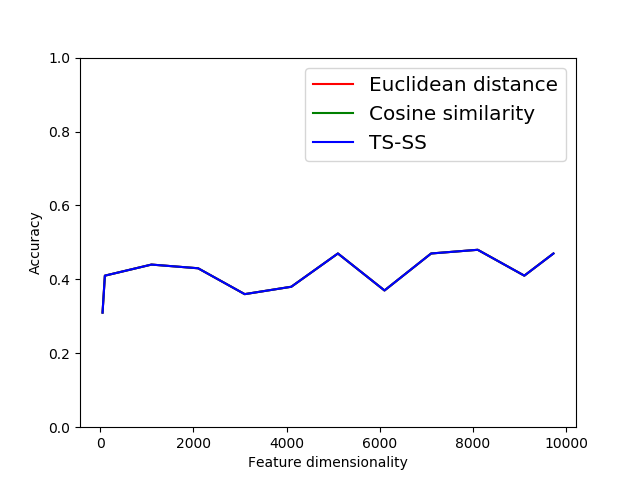}}
\caption{Predictive performances over the Brown dataset due to l2 (\ref{eq:l2}) constraint.}
\label{fig:brown_l2}
\end{figure}

\begin{figure}[h]
\centerline{\includegraphics[width=0.39\textwidth]{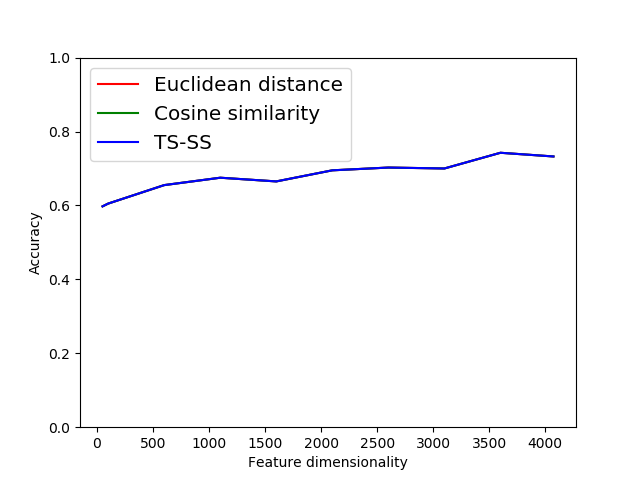}}
\caption{Predictive performances over the Movie Reviews dataset due to l2 (\ref{eq:l2}) constraint.}
\label{fig:movie_reviews_l2}
\end{figure}

\begin{figure}[h]
\centerline{\includegraphics[width=0.39\textwidth]{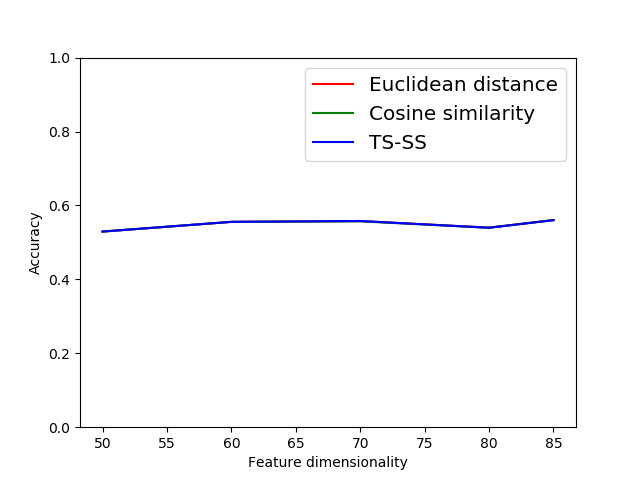}}
\caption{Predictive performances over the Sentence Polarity dataset due to l2 (\ref{eq:l2}) constraint.}
\label{fig:sentence_polarity_l2}
\end{figure}

\newpage
\section{}
\label{Appendix_l1_norm}

\begin{figure}[h]
\centerline{\includegraphics[width=0.39\textwidth]{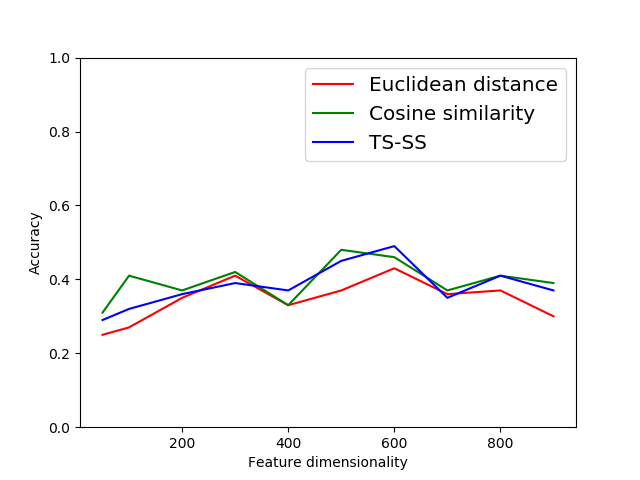}}
\caption{Predictive performances over the Brown dataset due to l1 (\ref{eq:l1}) constraint.}
\label{fig:brown_l1}
\end{figure}

\begin{figure}[h]
\centerline{\includegraphics[width=0.39\textwidth]{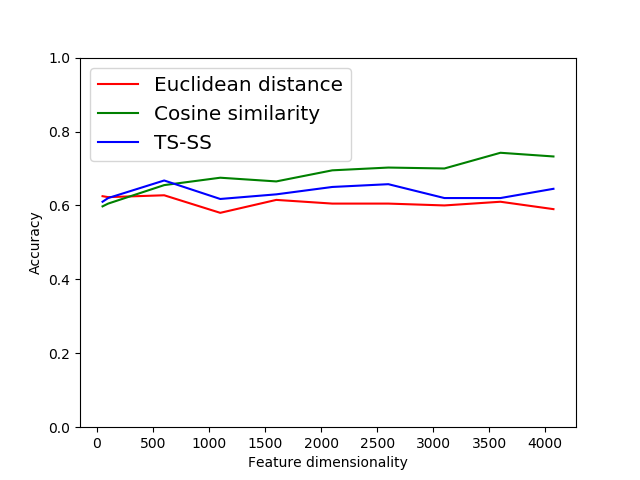}}
\caption{Predictive performances over the Movie Reviews dataset due to l1 (\ref{eq:l1}) constraint.}
\label{fig:movie_reviews_l1}
\end{figure}

\begin{figure}[h]
\centerline{\includegraphics[width=0.39\textwidth]{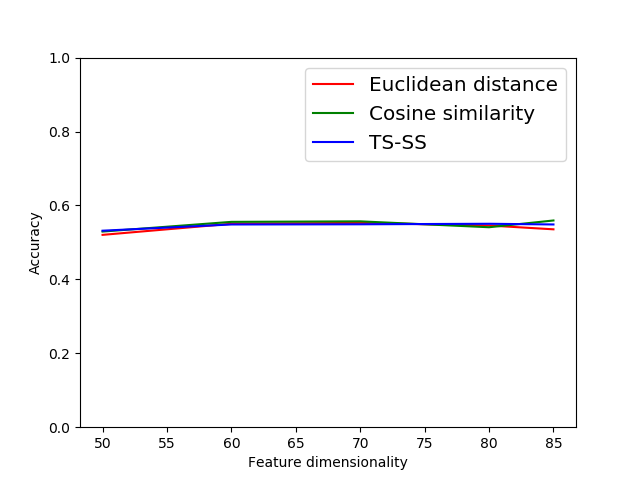}}
\caption{Predictive performances over the Sentence Polarity dataset due to l1 (\ref{eq:l1}) constraint.}
\label{fig:sentence_polarity_l1}
\end{figure}
% \subsection{First Subsection In Appendix}
% \label{FirstSubsectionAppendix}

% \section{Second Appendix}
% \label{SecondAppendix}

% \section{Third Appendix}
% \label{ThirdAppendix}

% \section{Fourth Appendix}
% \label{FourthAppendix}

\end{document}